\documentclass[runningheads]{llncs}
\makeatletter
\newcommand{\printfnsymbol}[1]{%
	\textsuperscript{\@fnsymbol{#1}}%
}
\usepackage{bbding}
\usepackage{graphicx}
\usepackage{amssymb}
\usepackage{amsmath}

\usepackage{multirow}
\usepackage{url}
\begin{document}

	\title{Scene-Aware Prompt for Multi-modal Dialogue Understanding and Generation\thanks{Supported by the National Key R\&D
			Program of China (2018YFB1305200), the National
			Natural Science Fund of China (62171183).}}
%	\author{Anonymous Author(s)}
	%
%	\authorrunning{Anonymous Author(s) et al.}
%	\institute{Anonymous}
		\author{Bin Li\inst{1}\thanks{These authors contribute this work equally.} \and
				Yixuan Weng\inst{2}\printfnsymbol{2} \and 
				Ziyu Ma\inst{1}\printfnsymbol{2}\and
				Bin Sun\inst{1} \and
				Shutao Li \inst{1}
			}
			\renewcommand{\lastandname}{\unskip,}
	%	%
		\authorrunning{Li et al.}
		\institute{College of Electrical and Information Engineering, Hunan University \\
				\email{\{libincn, maziyu, shutao\_li, sunbin611\}@hnu.edu.cn} \and
				{National Laboratory of Pattern Recognition Institute of Automation, \\
						Chinese Academy Sciences, Beijing} \\
				\email{wengsyx@gmail.com}}
	% 	}
%	%
\titlerunning{Scene-aware Prompt for MDUG}
%	\authorrunning{F. Author et al.}
%	% First names are abbreviated in the running head.
%	% If there are more than two authors, 'et al.' is used.
%	%
%	\institute{Princeton University, Princeton NJ 08544, USA \and
	%		Springer Heidelberg, Tiergartenstr. 17, 69121 Heidelberg, Germany
	%		\email{lncs@springer.com}\\
	%		\url{http://www.springer.com/gp/computer-science/lncs} \and
	%		ABC Institute, Rupert-Karls-University Heidelberg, Heidelberg, Germany\\
	%		\email{\{abc,lncs\}@uni-heidelberg.de}}
%
\maketitle              % typeset the header of the contribution
\vspace{-0.45cm}
\begin{abstract}
	This paper introduces the schemes of Team LingJing’s experiments in NLPCC-2022-Shared-Task-4 Multi-modal Dialogue Understanding and Generation (MDUG). The MDUG task can be divided into two phases: multi-modal context understanding and response generation. To fully leverage the visual information for both scene understanding and dialogue generation, we propose the scene-aware prompt for the MDUG task. Specifically, we utilize the multi-tasking strategy for jointly modelling the scene- and session- multi-modal understanding. The visual captions are adopted to aware the scene information, while the fixed-type templated prompt based on the scene- and session-aware labels are used to further improve the dialogue generation performance. Extensive experimental results show that the proposed method has achieved state-of-the-art (SOTA) performance compared with other competitive methods, where we rank the 1-st in all three subtasks in this MDUG competition.
\end{abstract}
\section{Introduction}
With advances in Al technology, the researchers are constructing intelligent machines capable of communicating with humans towards a given task \cite{de2017guesswhat}. The commonest option for achieving human-robot interaction is the design of a dialogue system that acts as a voice-interactive interface between the user and robot for better human-robot relationship \cite{deldjoo2021towards}. It is consequently becoming more and more important to equip systems with this social capability, with which they can respond appropriately to the user. Considering both the contextual and content elements of a multi-modal dialogue, many attempts for modelling have been made to enhance the human-robot services. For the dialogue modelling, the conversational context has played an essential part in determining the relevance of responses to a user's discourse \cite{StephenRoller2021RecipesFB}. Situational factors include any information used to characterise a dialogue situation that can influence the system's response, such as the mood of the user \cite{zhou2018emotional} or the environmental cues of the dialogue \cite{su2022language}. By taking contextual information into account, dialogue systems can quickly and automatically accommodate changes in the environment in which they operate, resulting in better user experiences.  A number of researchers regard images as visual contexts and have begun to develop multi-modal learning models to integrate images (scene) and text (sentences) \cite{Wang_NLPCC-2022-Shared-Task-4_2022}. Efforts on such integration can be categorized into two types: capturing (or summarising) the image presented or answering questions related to the content of the image provided. The former focuses on extracting image features as a basis for generating text, while the latter is the task of generating textual answers to textual questions about multimedia content, which is also called visual question answering (VQA) \cite{antol2015vqa}. For instance, Xu \textit{et al}. present the use of attention for image illustration, where image perception can be enhanced \cite{xu2015show}. Zhu \textit{et al}. further extend the application of spatial attention to the QA model \cite{zhu2016visual7w}. \par
The VQA research described above has been further broadened to include research on Audio-Visual Scene Sensing Dialogue (AVSD) \cite{alamri2019audio}, which aims to answer questions based on video clips. In order to achieve this goal, the system needs to properly combine different types of information extracted from the video in order to produce correct textual answers. In AVSD, the first few rounds of the dialogue are considered as extra-textual knowledge, forming a special context to improve the performance of the dialogue. As can be seen, conducting video-based conversations can present additional complexity and challenges \cite{Wang_NLPCC-2022-Shared-Task-4_2022}. Extracting features from the video not only deals with the inherent complexity of extracting image features, but also with the temporal interactions between image frames \cite{fang2021clip2video}. Besides, learning useful features becomes more difficult due to the limited availability of visual data \cite{li2022towards}.
\par
To further investigate the above challenges, NLPCC-2022-Shared-Task-4 designed the Multi-modal Dialogue Understanding and Generation (MDUG) task. This task aims at generating responses that are coherent to the dialogue and relevant to the video context.  In this paper, we present the scene-aware prompt method for the MDUG task.  The multi-task objectives are designed to jointly optimize the scene- and session- sequence prediction task. The visual captions are utilized to percept the scene information of the video, while the fixed-type templated prompt based on the scene- and session-aware labels are used to enhance dialogue generation. Extensive experimental results show that the proposed method has achieved state-of-the-art (SOTA) performance compared with other competitive baselines, where we rank the first in all three subtasks in this MDUG competition.
\par
Our main contributions are three-fold:
\vspace{-0.25cm}
\begin{itemize}
	\item We formulate the multi-modal understanding task as the joint multi-task modelling for better scene and session learning.
	\item To better leverage the scene information, we design the scene-aware prompt for the MDUG task, where the visual captions and fixed-type templated prompt with the scene- and session-aware labels are used to further improve the dialogue generation performance.
	\item Extensive experimental results show that the proposed method has achieved state-of-the-art (SOTA) performance compared with other competitive methods, which demonstrates its effectiveness.
\end{itemize}
\section{Task Introduction}
%\vspace{-0.2cm}
\subsection{Problem Definition}
This multi-modal dialogue understanding and generation task includes three tracks:
\begin{enumerate}
	\item 	Dialogue scene identification: predict the boundaries of different dialogue scenes given a set of continuous dialogue utterances and a related video.
	\item 	Dialogue session identification: predict the boundaries of different dialogue sessions given a set of continuous dialogue utterances and a related video (which is identical to Track 1).
	\item 	Dialogue response generation: generate a response based on scene and session predictions, while coherently catching up with the conversation.
\end{enumerate}
For pursuing these tasks, we formulated these tasks as follows, where the notation $V$ represents the video clips, and notation $C$ is the input dialogue context.
\begin{enumerate}
	\item 	For dialogue scene identification and dialogue session identification tasks, the final prediction (i.e., 0 or 1) is obtained through the input of $V$ and $C$, where we adopt a multi-task end-end framework to jointly perform these tasks.
	\item 	For dialogue response generation, given the scene ($S_{i}$) and session ($T_{i}$) predictions, the final response is generated with the pre-trained language model, where the clip captioners and the identified labels used as the prompt for providing extra context knowledge.
\end{enumerate}
\subsection{Evaluation Metric}
For the dialogue scene identification and dialogue topic identification tracks, we mainly use the accuracy metric (i.e.,  Acc$_s$ and Acc$_t$) for the final evaluation ranking. The calculation equation is shown as follows.
\begin{equation}
	\text{Acc}_{s}=\frac{1}{n} \sum_{i=1}^{n}l_{\left\{S_{i}=S_{i}{}^{\prime}\right\}}
\end{equation}
where $l$ is the sample number, the S$_i$ represents each predicted sample, and S$_i^{\prime}$ is the ground truth label. Similarly, we can present the Acc$_t$ below.	
\begin{equation}
	\text{Acc}_{t}=\frac{1}{n} \sum_{i=1}^{n}l_{\left\{T_{i}=T_{i}{}^{\prime}\right\}}
\end{equation}
\par Also, the F1 metric is adopted for reference, where the definitions of F1 are shown as follows:

\begin{equation}
	\text { F1 }=\frac{2 \times \text { Acc } \times \text { Recall }}{\text { Acc }+\text { Recall }}
\end{equation}

\par For the track 3, we adopt the BLEU \cite{papineni2002bleu}, ROUGE \cite{lin2004rouge}, METEOR \cite{banerjee2005meteor}, and CIDER \cite{vedantam2015cider} scores of the generated response for further evaluations\footnote{\url{https://github.com/tylin/coco-caption}}.

\subsection{Dateset}
% \begin{table}[t]
	% 	\label{datasets11}
	% 	\centering \small
	% 	\renewcommand\arraystretch{1.2}	\setlength{\tabcolsep}{7.7mm}
	% 	\begin{tabular}{llll}
		% 		\noalign{\hrule height 1pt}
		% 		Item & Train & Valid & Test \\ \hline
		% 		Clips & 40,006 & 1,955 & 1,934 \\ \hline
		% 		Utterances & 1,000,079 & 50,032 & 50,131 \\ \hline
		% 		Scenes & 56,535 & 3,202 & 3,284 \\ \hline
		% 		Sessions & 106,078 & 6,331 & 6,949 \\ \hline
		% 		Utter/clip & 25 & 25.6 & 25.92 \\ \hline
		% 		Scene/clip & 1.41 & 1.64 & 1.7 \\ \hline
		% 		Session/clip & 2.65 & 3.24 & 3.59 \\ \hline
		% 		En\_word/clip & 166.46 & 174.49 & 178.65 \\ \hline
		% 		En\_word/utter & 6.66 & 6.82 & 6.89 \\ \hline
		% 		Ch\_word/clip & 267.74 & 283.7 & 286.42 \\ \hline
		% 		Ch\_word/utter & 10.71 & 11.09 & 11.05 \\ 	
		% 		\noalign{\hrule height 1pt}
		% 	\end{tabular}
	
	% 	\caption{Details of the datasets in NLPCC shared task 4.}
	% \end{table}
The NLPCC shared task 4 presents three shared tasks \cite{Wang_NLPCC-2022-Shared-Task-4_2022}, namely, dialogue scene identification, dialogue session identification, and dialogue response generation. The ultimate goal is to generate a response that is coherent to the dialogue context and relevant to the video context.  \par	
%This competition As shown in the Table \ref{datasets11}, it is composed of
The dataset of the competition\footnote{\url{https://github.com/patrick-tssn/NLPCC-2022-Shared-Task-4}} contains 40,006, 1,955 and 1,934 video clips as the visual context. The dialogue contains 1,000,079, 50,032 and 50,131 utterances in the train, dev and test sets respectively. The source of the videos and dialogues for this task are crawled from online American TV series, which are split into the training, validation, and test sets. Each sample contains a series of dialogue utterances, which is associated with the video clip (downsampled to 3fps) during the dialogue duration. Each clip is processed in the ``jpg'' format for further modeling.  \par

\vspace{-0.3cm}
% modeling 
\section{Main Methods}
In this section, we will introduce our method in the three shared tasks in the NLPCC tasks 4, including multi-tasking multi-modal dialogue understanding and scene-aware prompt multi-modal dialogue generation. 
\vspace{-0.4cm}
\subsection{Multi-tasking Multi-modal Dialogue Understanding}
\begin{table}[t]
	\centering \small
	\renewcommand\arraystretch{1.2}	\setlength{\tabcolsep}{4.7mm}
	\quad
	\begin{tabular}{ccc}
		\noalign{\hrule height 1pt}
		Scene Label & Session Label & Co-occurrence \\
		\hline
		1     & 0       & \XSolidBrush          \\
		0     & 1       &  \CheckmarkBold              \\
		1     & 1       &  \CheckmarkBold              \\
		0     & 0       & \CheckmarkBold   \\           
		\noalign{\hrule height 1pt}
	\end{tabular}
	\label{co-oc}
	\caption{Co-occurrence of the labels in both shared tasks 1 and 2.}
	\vspace{-0.7cm}
\end{table}
Multi-modal dialogue understanding is still a great challenge since the visual and textual modalities share different information \cite{de2017guesswhat}. Both shared task 1 and shared task 2 share the same task objectives, which is to perform sequence prediction for the multi-modal dialogue understanding. Considering the intrinsic co-occurrence between the two labels (i.e., scene and session labels), we can obtain Table 1. From this table, we can conclude that both tracks share some common label information about the co-occurrence relationship. As a result, we propose the multi-tasking method for jointly training both tracks.

\begin{figure*}[t]
	\centering
	\includegraphics[scale=0.47]{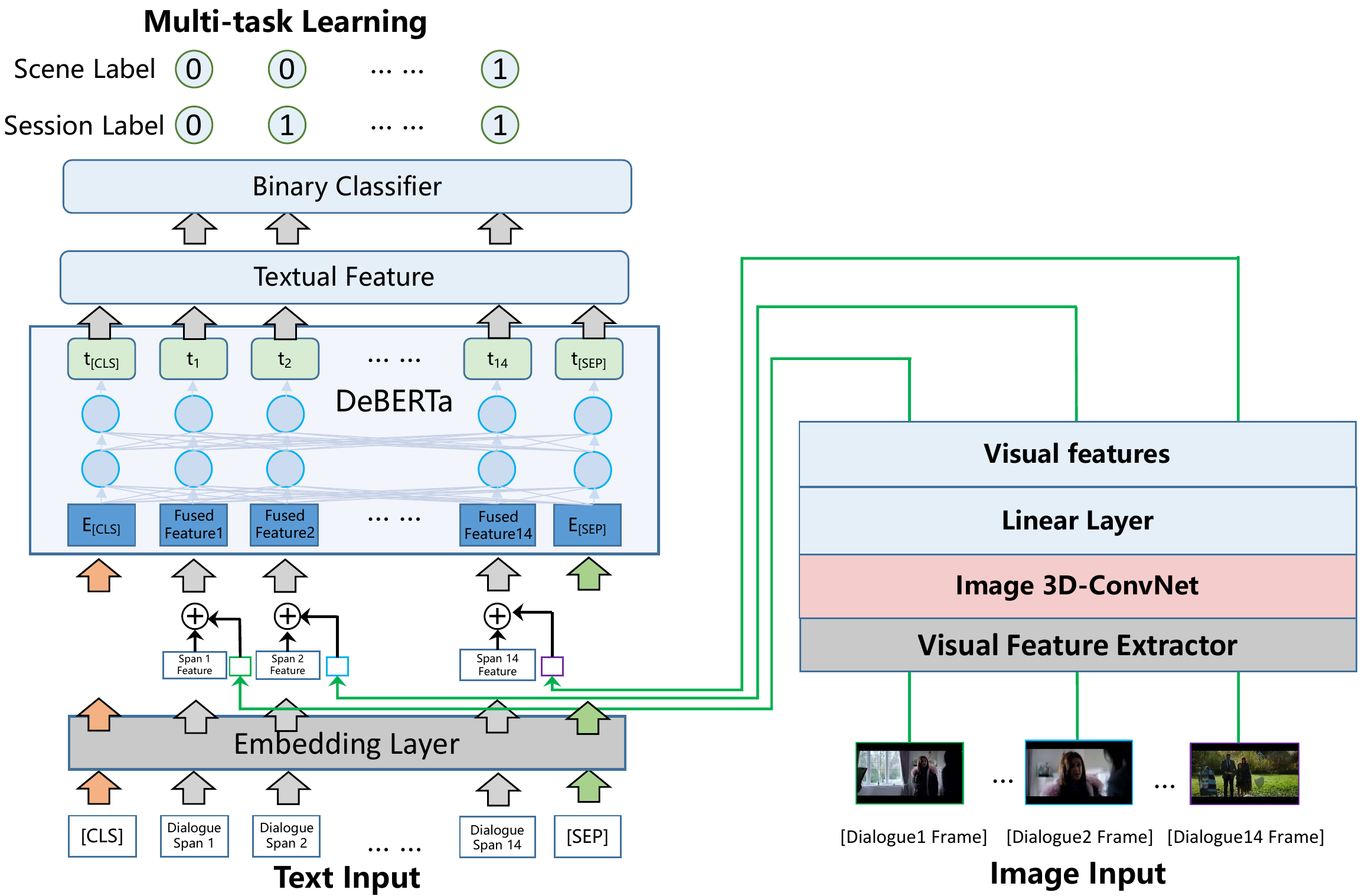}
	\caption{Overview of the proposed multi-tasking multi-modal dialogue understanding.}
	\label{taskdes1112}
	\vspace{-0.6cm}
\end{figure*}

%我们为了促进深层次的多模态信息融合，我们首先使用I3D模型对视频特征进行提取与处理，在这之后，我们对原有的语言模型结构进行更改，我们利用时间线进行文本与视觉的逐帧对齐。具体来说，我们首先将每一句话和当前视频帧进行融合，我们使用文本字幕的Token的Embedding层输出，找到相同时间点的图像，进行点乘融合。我们利用相同时间点的信息特征能够使得多模态交互更加精细与完整。
As shown in the Figure \ref{taskdes1112}, we use the I3D model \cite{JoaoCarreira2017QuoVA} to extract and process video features to promote deep multi-modal information fusion. After that, we change the structure of the original language model. Specifically, we use the timeline to align the dialogue span and visual frame, where each input text span is fused with the current video frame. The embedding layer can produce the fused features by adding fusion with the output text span and the corresponded frame. This fusion strategy can make the multi-modal interaction more precise and efficient \cite{sun2021non}.

For the final multi-task modelling, we design a linear layer for the output layer in each task. Specifically, the \textit{[SEP]} feature vector of each sentence can produce the output with the two different binary linear layers. The multi-task modelling is designed to perform the scene and session sequence prediction, where two prediction results are obtained meanwhile for each input.
%然后，我们对预训练语言模型的输出层中加入了一个线性层，我们取得每一句话的[SEP]特征向量，输入到两个不同的二分类线性层中，一个负责预测scene信息，另一个负责预测session信息。每一句需要判断的字幕都是有且仅有一个预测结果。

\vspace{-0.4cm}
\subsection{Scene-aware Prompt Multi-modal Dialogue Generation} 
Traditional text-based generation methods have limitations in multi-modal scenarios \cite{deldjoo2021towards}. On the one hand, it is because the interaction of characters in a multi-modal scene not only relies on the textual information of the dialogue context, but also needs to depend on the prompts of the environment scene and dialogue session \cite{su2022language}. On the other hand, the ability of a single text modality to perceive the multi-modal dialogue context is limited, and it is a wise choice to augment and enrich the dialogue context with other modalities \cite{li2022vpai_lab}. \par
\begin{figure*}[t]
	\centering
	\includegraphics[scale=0.405]{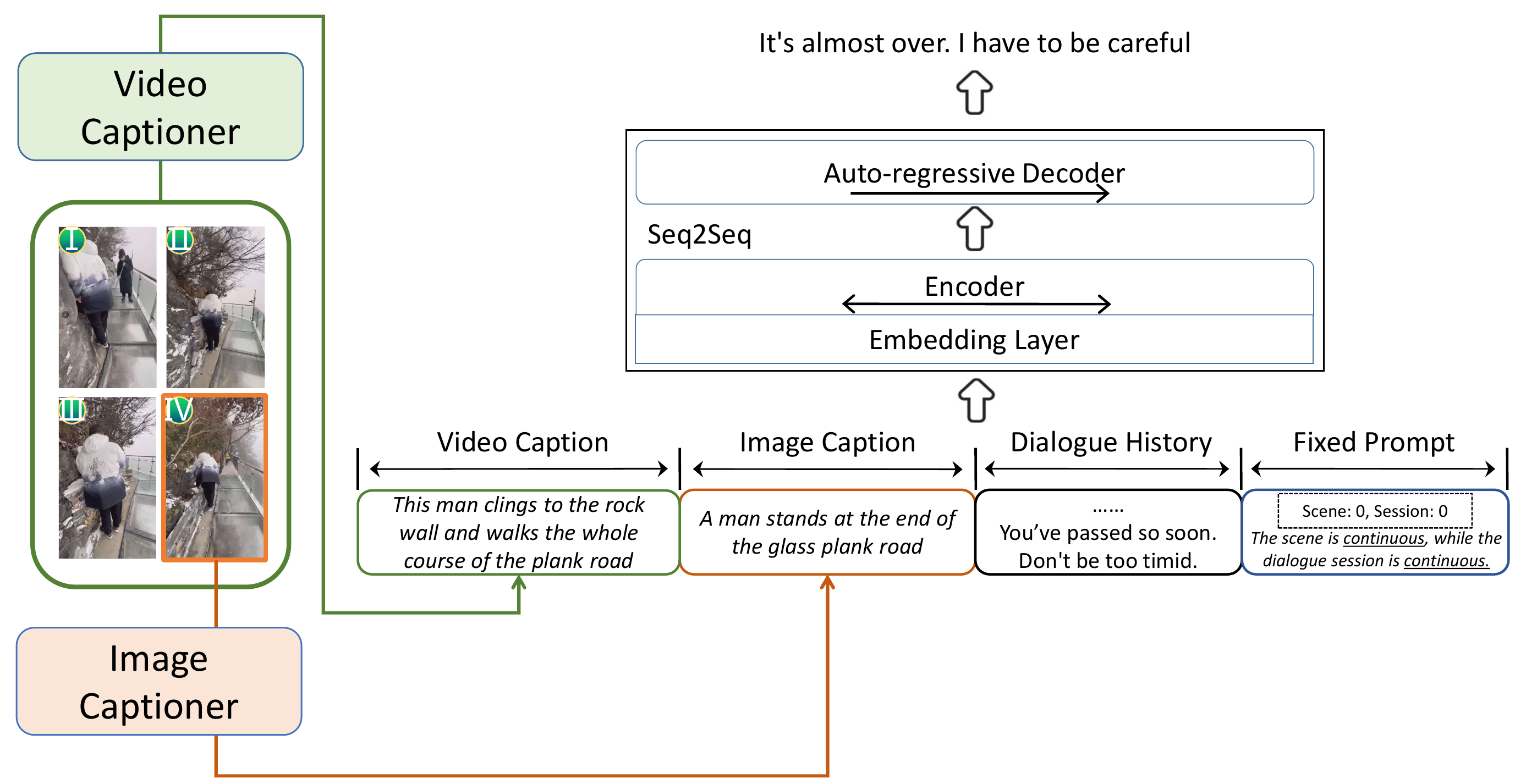}
	\caption{Overview of the proposed scene-aware prompt multi-modal dialogue generation.}
	\label{taskdes11}
	\vspace{-0.7cm}
\end{figure*}
Therefore, we propose a multi-modal dialogue generation method based on the scene-aware prompt, which is shown in the Figure \ref{taskdes11}. Specifically, we use the pre-trained video captioner model (i.e., UniVL \cite{luo2020univl}) and image captioner model (i.e., BLIP \cite{li2022blip}) to obtain the caption information of the environmental scene, which further enhances the multi-modal dialogue generation. At the same time, we design a fixed-type templated prompt based on the scene- and session-aware labels to further improve the controllability of the generated responses.
\vspace{-0.3cm}
\subsubsection{Visual Caption}
The visual caption contains two types of information, which are the video caption and the image caption. For the video caption, we adopt the UniVL pre-trained model as the scene-aware information extractor, which is the state-of-the-art (SOTA) video captioner. This model is used for extracted the video captions from the video clips during each utterance. Also, we consider that the last frame for the dialogue response also contains further information for the dialogue response generation. As a result, we adopt the BLIP pre-trained model for extracting the last clip frame to obtain the image captions. Since the caption both from the video crawled from the American online TV shows, we use these models with zero-shot for this dialogue caption generation task  \cite{wang2022language}. Both the above step can be represented as follows.
\begin{equation}
	\begin{aligned}
		\mathbf{{T_{video}}}&=\phi(\mathcal{V}_{\mathrm{video clips}})\\
		\mathbf{{T_{image}}}&=\varphi(\mathcal{V}_{\mathrm{image clip}})
	\end{aligned}
\end{equation}
where ${T}\in\mathbb{R}^{d}$, $\mathcal{V}_{\mathrm{image}}\in\mathbb{R}^{k}$,d is the dimension, which is the same as text predictor encoder embedding. The $\phi$ and $\varphi$ represent the model parameter of the video captioner and image captioner respectively. 
\vspace{-0.4cm}
\subsubsection{Prompt Design} 
After obtaining the scene- and session- predicted labels, we shall utilize this information to provide extra knowledge from the scene environment. Specifically, we design the fix-type prompt for the pre-trained language model, where the prompt is used as the input text tokens concatenated with the visual captions and the dialogue contexts. On the one side, the fix-typed prompt covers the information from the visual clips. On the other side, the prompt can be well-formed information that controls the dialogue text generation \cite{liu2021pre}. The fix-typed prompt is designed as ``The scene is \underline{continuous}, while the dialogue session is \underline{not continuous}'', where the scene and session labels are 0 and 1 respectively.
\vspace{-0.3cm}
\subsubsection{Prompt Tuning} 
Intuitively, the dialogue contexts $\mathbf{C}$, visual captions $\mathbf{{T_{video}}}$, $\mathbf{{T_{image}}}$ and the scene-aware prompt $\mathbf{{T_{prompt}}}$ are used as the input tokens which are concatenated together. The [CLS] is positioned at the head of the input token, while the rest text tokens are sent to the embedding layer ($\mathbf{Emb}$) as the trigger to model to generate the response. The above process is presented as follows
\begin{equation}
	\small
	%			\hspace{-0.31cm}
	\mathbf{Input}=\mathbf{Emb}\left([\mathrm{CLS}]+ {\mathbf{C}}+ [\mathrm{SEP}] + \mathbf{{T_{video}}} + [\mathrm{SEP}] + \mathbf{{T_{video}}} + [\mathrm{SEP}] + \mathbf{{T_{prompt}}}  \right)
\end{equation}

\par After concatenation, the embedding module is adopted for learning the features in the same vector space for the further response generation. 
%Unlike precious image models jointly train an image feature extractor and a linear classifier to predict some label, CLIP jointly trains an image encoder and a text encoder to .
% 	\subsubsection{Hierarchical Decomposition Positional Embedding} 
% 	%% 层次分解位置编码
\vspace{-0.4cm}
\subsubsection{Response Generation} Finally we generate response with seq2seq model's decoder. We define $L_{R}$ as the auto-regressive decoder loss. 
\begin{equation}
	\mathcal L_{R}(\psi) =-\sum_{t} \log P_{\varphi}\left(y_{t} \mid y_{0}, \ldots, y_{t-1},\mathbf{Input} \right)
	%			&=-\sum_{t} \log P_{\varphi}\left(y_{t} \mid O_{F}\right)
\end{equation}
where notation $\psi$ represents the parameters of the pre-traned model. The $i$ represents the $i$-th word generated by the decoder, $y_0,\ldots , y_{t-1}$ is the generated tokens, and $y_t$ is the next token.
\vspace{-0.4cm}
\subsection{Training and Inference}
For the training step, each sample is concatenated with the visual captions and the fixed prompt, where the visual captions are generated by the zero-shot video and image captioners, and the fixed prompt is produced by the corresponded scene and session labels.
\par
For inference, we first predict the scene and session labels of the last turn, and then we translate the predicted labels into the fixed prompt for further concatenation. Also, the video clips and last image frame are implemented for obtaining the visual captions. Finally, the dialogue context, visual captions and the fixed prompt are concatenated for the response generation.
\vspace{-0.4cm}
\section{Experiments}
In this section, we describe the specific implementation steps of the experiment and show the experimental results of our method in the MDUG dataset.
%% MDUG
\vspace{-0.4cm}
\subsection{Experimental Setup}

We conduct different experiments in three tasks in the MDUG dataset. Specifically, we use some pre-trained language models as the baseline methods. In subtask 1/2, we use the BERT\footnote{\url{https://huggingface.co/bert-base-uncased}} \cite{devlin2018bert}  which a language mask model (MLM) \cite{WilsonLTaylor1953CLOZEPA} task pre-training model. The RoBERTa\footnote{\url{https://huggingface.co/roberta-large}} \cite{YinhanLiu2019RoBERTaAR} has conducted MLM pre-training for a long period time. The ELECTRA\footnote{\url{https://huggingface.co/google/electra-large-discriminator}} \cite{KevinClark2020ELECTRAPT} uses the replaced token detection task instead of the MLM task to obtain higher training efficiency. The DeBERTa-v3 \cite{PengchengHe2021DeBERTaV3ID} implements gradient decoupling on the basis of Electra to avoid the tug-of-war procedure \cite{RaiaHadsell2020EmbracingCC}.

In subtask 3, we select some strong baseline models in the generation domain for comparison. The BART \cite{MichaelLewis2019BARTDS} and the T5 \cite{2020t5} use the denoising and mask restoration task for pre-training respectively. The BART has achieved SOTA in generation tasks like translation, while T5 has SOTA performance in understanding and summarization.

In order to get a strong baseline, we use DeBERTa-v3-large as the backbone network in the understanding task (subtask 1 and 2) and Blender \cite{StephenRoller2021RecipesFB} as the backbone network in the generation task (subtask3). 

All the hyperparameters are adjusted in the dev set to ensure fairness. In all our experiments, at the end of each epoch of training, we will test in the development dataset, and select the highest model (mainly depending on Acc or BLEU) to predict in the test dataset. All the tables report the highest score in the development set except for the final official score table. All the experimental results are repeated three times, and the highest and lowest scores are removed. 

We set the maximum token length to 512 and delete the excess text. We have fine-tuned 10 epochs of training in three A100 GPUs on the Pytorch\footnote{\url{https://pytorch.org}} and the hugging-face\footnote{\url{https://github.com/huggingface/transformers}} framework, with a batch size of 10. We implement distributed training with mixed precision based on the DeepSpeed \cite{10.5555/3433701.3433727}. We use an AdamW optimizer \cite{IlyaLoshchilov2018DecoupledWD} with a maximum learning rate of $1 \times 10^{-5}$, followed by linear attenuation and warm-up optimizing schedules \cite{7780459}.

\begin{table*}[t]
	\centering
	
	\quad
	\begin{minipage}[]{1\textwidth}
		\centering
		\resizebox{\textwidth}{!}{%
			\begin{tabular}{l|cccc}
				\hline
				\multicolumn{1}{c|}{\textbf{Models}}                        &\ \ \ \ \ \textbf{Acc} \ \ \ \ \ & \ \ \ \ \ \textbf{F1}\ \ \ \ \ &\ \ \ \ \ \textbf{Precision}\ \ \ \ \ &\ \ \ \ \ \textbf{Recall} \ \ \ \ \ \   \\\hline
				\multicolumn{1}{l|}{\textbf{Random Mode}}                      &   49.653&11.308&6.369&\textbf{50.363} \\ \hline
				\textbf{BERT-base}\cite{MichaelLewis2019BARTDS} (2019) & {91.627}&0&0&0\\
				\textbf{RoBERTa-large}\cite{YinhanLiu2019RoBERTaAR}(2019) &91.432&15.683&21.030&\underline{12.504}\\
				\textbf{ELECTRA-large}\cite{KevinClark2020ELECTRAPT}(2020) &92.383&16.316&27.212&11.651 \\
				\textbf{DeBERTa-v3-large}\cite{PengchengHe2021DeBERTaV3ID}(2021) &92.961&17.848&34.830&11.998 \\ \hline
				
				\textbf{Ours (Single-task)} &  \underline{93.567}&\underline{19.329}&\underline{48.116}&12.093 \\
				\textbf{Ours (Multi-task)} & \textbf{93.794} &\textbf{19.854}&\textbf{56.094}&12.062 \\ \hline
				
			\end{tabular}%
		}
		\caption{Performance comparison of the variants methods on MDUG dataset  for subtask 1. We highlight the best score in each column in \textbf{bold}, and the second-best score with \underline{underline}. We also show improvements between first place and second place.
			% The best performing (in terms of IoU=0.7) FPL value is considered to experiment with \vslqgh{} model.
		}
		\label{3}
	\end{minipage}%
	\vspace{-0.7cm}
\end{table*}

\begin{table*}[t]
	\centering
	
	\quad
	\begin{minipage}[]{1\textwidth}
		\centering
		\resizebox{\textwidth}{!}{%
			\begin{tabular}{l|cccc}
				\hline
				\multicolumn{1}{c|}{\textbf{Methods}}                        &\ \ \ \ \ \textbf{Acc} \ \ \ \ \ & \ \ \ \ \ \textbf{F1}\ \ \ \ \ &\ \ \ \ \ \textbf{Precision}\ \ \ \ \ &\ \ \ \ \ \textbf{Recall} \ \ \ \ \ \   \\\hline
				\multicolumn{1}{l|}{\textbf{Random Mode}}                      &   49.534&19.919&12.455&\textbf{49.713} \\ \hline
				\textbf{BERT-base}\cite{MichaelLewis2019BARTDS} (2019) & 87.375&0&0&0\\
				\textbf{RoBERTa-large}\cite{YinhanLiu2019RoBERTaAR}(2019) &87.912&34.040&54.712&24.705\\
				\textbf{ELECTRA-large}\cite{KevinClark2020ELECTRAPT}(2020) &87.580&34.242&51.638&25.614 \\
				\textbf{DeBERTa-v3-large}\cite{PengchengHe2021DeBERTaV3ID}(2021) &87.912&35.038&54.490&\underline{25.821} \\ \hline
				
				\textbf{Ours (Single-task)} &  \underline{88.075}&\underline{35.078}&\underline{56.097}&25.518 \\
				\textbf{Ours (Multi-task)} & \textbf{88.248} &\textbf{35.484}&\textbf{57.811}&25.598 \\ \hline
				
			\end{tabular}%
		}
		\caption{Performance comparison of the variants methods on MDUG dataset for subtask 2. We highlight the best score in each column in \textbf{bold}, and the second-best score with \underline{underline}. We also show improvements between first place and second place.
			% The best performing (in terms of IoU=0.7) FPL value is considered to experiment with \vslqgh{} model.
		}
		\label{4}
	\end{minipage}%
	\vspace{-0.5cm}
\end{table*}

\begin{table*}[t]
	\centering
	\label{5}
	\quad
	\begin{minipage}[]{1\textwidth}
		\centering
		\label{54}
		\resizebox{\textwidth}{!}{%
			\begin{tabular}{l|cccc|c}
				\hline
				\multicolumn{1}{c|}{\textbf{Methods}}                        & \ \ \textbf{BLEU-1} \ \ & \ \ \textbf{ROUGE-L} \ \ &\ \ \textbf{METEOR}\ \  & \ \ \textbf{CIDEr} \ \ \ & \ \ \textbf{Avg} \ \ \\ \hline
				\multicolumn{1}{l|}{\textbf{Random Mode}}                      & 4.81&	3.92&	2.21 &	0.02   & 2.72  \\ \hline
				\multirow{1}{*}{\rotatebox[origin=c]{0}{\textbf{BART-base}\cite{MichaelLewis2019BARTDS} (2019)}}
				&  5.74   & 6.10            & 3.87             & 0.04 & 3.94
				
				\\ 
				\multirow{1}{*}{\rotatebox[origin=c]{0}{\textbf{T5-base}  (2020)}}
				&  2.94   & 4.44            & 2.81   & 0.01             & 2.55       \\ 
				
				\multirow{1}{*}{\rotatebox[origin=c]{0}{\textbf{Blender-400M \cite{StephenRoller2021RecipesFB}}(2021)}}
				
				&  7.01   &  8.73           & 6.05             & 0.06  &5.46     \\ 
				\hline
				\multirow{1}{*}{\rotatebox[origin=c]{0}{\textbf{Ours (Single-task)}}}
				
				&  \underline{11.9}   &  \underline{18.1}           & \underline{11.7}             & 0.57  &10.96     \\ 
				\multirow{1}{*}{\rotatebox[origin=c]{0}{\textbf{Ours (W/O Image Prompt)}}}
				
				&  10.8   &  17.5           & 8.2             & 0.84  &9.52     \\ 
				\multirow{1}{*}{\rotatebox[origin=c]{0}{\textbf{Ours (W/O Video Prompt)}}}
				
				&  12.9   &  18.7           & 9.1             & \underline{0.96}  &10.42     \\ 
				\multirow{1}{*}{\rotatebox[origin=c]{0}{\textbf{Ours (W/O Prompt)}}}
				
				&  8.7   &  15.5           & 7.6             & 0.78  &8.27     \\ 
				\multirow{1}{*}{\rotatebox[origin=c]{0}{\textbf{Ours (Multi-task)}}}
				
				&  \textbf{14.2}   &  \textbf{22.5}           & \textbf{12.1}             & \textbf{1.19}  &\textbf{12.47}     \\ 
				\hline
				
			\end{tabular}%
		}
		\caption{Performance comparison of the variants methods on MDUG dataset for subtask 3. We highlight the best score in each column in \textbf{bold}, and the second-best score with \underline{underline}. We also show improvements between first place and second place.
			% The best performing (in terms of IoU=0.7) FPL value is considered to experiment with \vslqgh{} model.
		}
		
	\end{minipage}%
	
	\vspace{-0.6cm}
\end{table*}
	\vspace{-0.2cm}
\subsection{Main Results}
% 首先我们进行了子任务1和2的实验。我们分别比较了预训练语言模型的性能和随机选择的性能。首先，我们可以看到Random Mode情况下的准确率只有不到50%，但拥有较高的召回率。这是因为我们发现，MDUG数据集的标签分布并不平衡。对于BERT的F1性能差，我们认为是由于标签极端不平衡导致的BERT无法拟合。我们检查发现BERT的输出结果全部选择0。

%我们的方法，相比起单一模态的预训练模型能够对视觉信息进行更深层次的感知与理解。在Table3和Table4中我们可以发现：对于Task1，我们的准确率和F1指标，分别能够提升0.833，2.006；对于Task2，我们的准确率和F1指标能够分别提升0.336和0.446.

%对于Task3任务，这是一个生成任务，因此我们选择了一些生成的基线模型进行对比。我们从中可以发现，我们的模型能够具有较大的提升。我们认为这是因为我们的方法能够在利用预训练语言模型的基础上，充分利用多模态特征信息。
We conducted experiments on subtasks 1 and 2, which are shown in Table \ref{3} and \ref{4} respectively. We compare the performance of the pre-trained language model with random selection. From the table, we can see that the accuracy rate in random mode is less than 50\%, but it has a high recall rate. This is because the label distribution of the MDUG dataset is not balanced. For the poor F1 performance of BERT, we consider that it is due to the extreme imbalance of labels so that the BERT cannot overfit these datasets, where the output recall and f1 results of the BERT are all 0. With the expansion of the pre-training scale, the effect of the model gradually improves. Compared with single-task modeling, our method with multi-tasking modeling can perceive and understand the visual information at a deeper level. Compared with other competitive methods, we can find that for subtask1, our accuracy and F1 index can be improved by 0.833\% and 2.006\% respectively. For subtask2, our accuracy and F1 index can be improved by 0.336\% and 0.446\% respectively.
\par

The subtask 3 is a generation task, so we selected some generated baseline models for comparison, which is shown in the Table 4. We can find that our model can be greatly improved, which is because our method can make full use of multi-modal feature information on the basis of the pre-training language model. Compared with other competitive methods, our method improves the average score by 7.01, which proves the effectiveness of our method.

\begin{table*}[t]
	\centering
	\label{6}
	\quad
	\begin{minipage}[]{1\textwidth}
		\centering
		\resizebox{\textwidth}{!}{%
			\begin{tabular}{c|c|c|cccc}
				\hline
				\ \ \ \ \ \ \ \ \textbf{Item} \ \ \ \ \ \ \ \ &\ \ \ \ \ \ \ \ \textbf{Objective} \ \ \ \ \ \ \ \ & \ \ \ \ \ \ \ \ \textbf{Rank} \ \ \ \ \  \ \ \                    & \ \ \ \ \ \ \ \ \textbf{Acc} \ \ \ \ \ \ \ \ & \ \ \ \ \ \ \ \ \textbf{F1} \ \ \ \ \ \ \ \ \\\hline
				
				\textbf{Subtask 1} & \textbf{Scene} &1& \textbf{93.88}&18.18 \\
				\textbf{Subtask 2}& \textbf{Session} &1& \textbf{87.79} &39.76 \\ \hline
				
			\end{tabular}%
		}
		\caption{The online result of the subtask 1 and 2.
			% The best performing (in terms of IoU=0.7) FPL value is considered to experiment with \vslqgh{} model.
		}
		
	\end{minipage}%
	\vspace{-0.1cm}
\end{table*}
\begin{table*}[t]
	\centering
	\label{7}
	\quad
	\begin{minipage}[]{1\textwidth}
		\centering
		\resizebox{\textwidth}{!}{%
			\begin{tabular}{c|c|cccc|c}
				\hline
				\multicolumn{1}{c|}{\textbf{Models}}                        &\textbf{Rank}& \ \ \ \ \ \textbf{BLEU-1} \ \ \ \ \ & \ \ \ \ \ \textbf{ROUGE-L}\ \ \ \ \ &\ \ \ \ \ \textbf{METEOR}\ \ \ \ \ &\ \ \ \ \ \textbf{CIDEr} \ \ \ \ \ \ & \ \ \ \ \ \textbf{Avg} \ \ \ \ \ \\ \hline
				
				\multirow{1}{*}{\rotatebox[origin=c]{0}{\textbf{Ours}}}& 1
				&  \textbf{13.9}   &  \textbf{22.6}           & \textbf{11.7}             & \textbf{1.29}  &\textbf{6.91}     \\ 
				\hline
				
			\end{tabular}%
		}
		\caption{The online result of the subtask 3.
			% The best performing (in terms of IoU=0.7) FPL value is considered to experiment with \vslqgh{} model.
		}
		
	\end{minipage}%
	\vspace{-0.5cm}
\end{table*}	
\vspace{-0.4cm}
\subsection{Ablation Study}

% 我们对实验进行了更深入的分析。我们测试了取消Co-learning的效果在Subtask1 和3中。我们发现，取消Separate Modeling之后，效果会显著下降。我们认为这是由于Scene和Session之间具有较大的相关性，通过多任务学习能够使得模型更好地理解文本与视觉信息的实际含义。这有助于模型增加泛化能力，提升最终效果。

%我们在Subtask3中进行了消融实验。我们通过取消原有的Image Prompt或者Video Prompt，使得模型缺乏一部分的视觉感知能力，以此来评估不同方法的有效性。在我们的实验中可以发现，模型的性能下降幅度并不明显如果缺乏任意一个视觉提示。但如果缺乏所有的视觉提示，这将导致模型缺乏视觉场景建模能力，从而大幅度影响模型的预测性能。
We also implement the ablation study for the proposed method, which is shown in Table 2 and 3. The image and video captions can improve the final results, and the prompt can enhance the performance of the dialogue generation. Also, with the aid of multi-tasking modeling, our method can be further improved.
\par
Specifically, we make a more in-depth analysis of the experiment. We test the effect of cancelling multi-task in subtasks 1 and 2. We find that the effect will decrease significantly after single-task modelling is cancelled. We consider it is a great correlation between scene and session. Through multi-task learning, the model can better understand the actual meaning of the textual and visual information. It helps to increase the generalization ability of the model and improve the final effect.

Moreover, we perform ablation experiments in subtask 3, which is shown in Table 4. We make the model without some visual perception ability by canceling the original image caption or video caption, so as to evaluate the effectiveness of different methods. In our ablation experiments, we can find that the performance degradation of the model is not obvious if any visual information is missing. However, if all visual cues are missing, the model will lack the ability of visual scene modeling, which will greatly affect the final prediction performance.
\vspace{-0.4cm}
\subsection{Online Results}
As for the online results, we reported the final results of our system. In Tabel 5 and 6, our system showed very convincing performance. We have achieved the first place in all subtasks, which fully demonstrates our method's effectiveness.
\vspace{-0.4cm}
\section{Conclusion}
\vspace{-0.2cm}
In this paper, it is mainly introduced that, in order to realize the better multi-modal dialogue understanding and generation, the LingJing team modelled the joint multi-task understanding tasks for subtasks 1 and 2.  In subtask 3, to better percept the scene information, we designed the scene-aware prompt method to leverage the visual information for the multi-modal dialogue generation. As a result, our team has won three subtasks in this MDUG completion, which demonstrates the effectiveness of the proposed method. However, there is still a long way for robust multi-modal understanding and generation, how to combine both capabilities well is yet to be explored.

\vspace{-0.2cm}

\bibliographystyle{unsrt}
\bibliography{a}
\end{document}